\pdfoutput=1

\documentclass[11pt, table]{article}
\usepackage{acl}
\usepackage{times}
\usepackage{latexsym}
\usepackage[T1]{fontenc}
\usepackage[utf8]{inputenc}
\usepackage{microtype}
\usepackage{inconsolata}
\usepackage{graphicx}
\usepackage[normalem]{ulem}
\usepackage{adjustbox}
\usepackage{float} 
\usepackage{natbib}
\usepackage{amssymb}
\usepackage{booktabs}
\usepackage{tabularx}
\usepackage{array}
\usepackage{amsmath}
\usepackage{makecell}
\usepackage{xspace}
\usepackage{multirow}

\newcommand{\llama}{\textsc{gmn-l}\xspace}
\newcommand{\swow}{\textsc{gmn-h}\xspace}
\newcommand{\gmsn}{\textsc{gmn}\xspace}

\newcommand{\wahuman}{\textsc{wa-h}\xspace}
\newcommand{\wallm}{\textsc{wa-l}\xspace}
\title{Comparing Moral Values in Western English-speaking societies and LLMs with Word Associations}

\author{Chaoyi Xiang$^1$ \quad Chunhua Liu$^1$ \quad Simon De Deyne$^2$ \quad Lea Frermann$^1$ \\
        $^1$School of Computing and Information Systems, 
        The University of Melbourne\\
        $^2$Complex Human Data Hub, 
        The University of Melbourne\\
        \texttt{chaoyix@student.unimelb.edu.au}\\
        \texttt{\{chunhua.liu1, simon.dedeyne, lea.frermann\}@unimelb.edu.au} % Grouped emails
}

\begin{document}
\maketitle

\begin{abstract}
As the impact of large language models increases, understanding the moral values they reflect becomes ever more important. Assessing the nature of moral values as understood by these models via direct prompting is challenging due to potential leakage of human norms into model training data, and their sensitivity to prompt formulation. Instead, we propose to use word associations, which have been shown to reflect moral reasoning in humans, as low-level underlying representations to obtain a more robust picture of LLMs' moral reasoning. We study moral differences in associations from western English-speaking communities and LLMs trained predominantly on English data. First, we create a large dataset of \textit{LLM-generated} word associations, resembling an existing data set of \textit{human} word associations. Next, we propose a novel method to propagate moral values based on seed words derived from Moral Foundation Theory through the human and LLM-generated association graphs. Finally, we compare the resulting moral conceptualizations, highlighting detailed but systematic differences between moral values emerging from English speakers and LLM associations.\footnote{All code and data are available at  \url{https://github.com/ChunhuaLiu596/Word_Association_Generation}}

\end{abstract}

\section{Introduction}

Large Language Models (LLMs) are trained on extensive corpora to learn linguistic patterns, contextual nuances, and implicit elements of human values. As these models are increasingly deployed in real-world applications, concerns have arisen regarding their moral alignment with humans~\cite{ji2024}. Assessing moral alignment poses a complex challenge because it remains unclear how to quantify an LLM's adherence to ethical principles and societal norms, given their next-token prediction nature~\cite{Scherrer2023} and their sensitivity to context and question framing, leading to varied responses~\cite{Almeida2024,nam2024,anagnostidis2024}. Moreover, the leakage of moral questionnaires into the LLMs' training data~\cite{abdulhai2023,dai2024} raises questions about the genuineness of their responses. 

\begin{figure}
\centering
    \includegraphics[width=\linewidth]{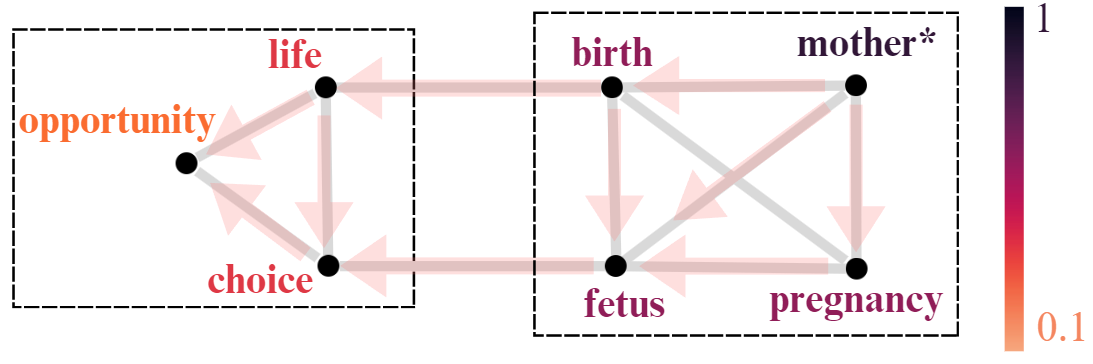}
\caption{An illustration of moral information propagation (colored nodes and arrows) through word associations (gray edges). Information is propagated from the moral seed word `mother' ($*$). The right box contains directly connected concepts with `mother', while the box on the left illustrates information flow to a more distant area in the graph. Color reflects the inferred moral intensity of a concept.
% , \textcolor{red}{with higher intensity indicating a stronger relevance to morality}.
}
\label{fig:p1}
\end{figure}

We present a framework for a more robust comparison of morality in humans and LLMs, focusing on moral values in western English-speaking cultures given their prevalence in prior research and LLM training data~\cite{henrich2010weirdest}. We address the limitations of existing methods that directly prompt LLMs with moral questionnaires, which have been shown to yield unreliable results~\cite{Almeida2024,Scherrer2023,abdulhai2023}. Instead, we measure the ``mental lexicon'' of LLMs using the well-established psychological paradigm of word associations~\cite{clark1970word,vanrensbergen2015}, see Figure~\ref{fig:p1}. In a typical word association experiment, participants are provided with a cue word and tasked with generating spontaneous associations. We pose the same task to LLMs to measure how LLMs internally organize and relate concepts. Previous work~\cite{Ramezani2024} has shown that moral values {of English language speakers} can be reliably recovered from {\it their} word associations. Here, we compare moral {values embedded} in {English} word associations from humans and LLMs, allowing for a more robust evaluation of LLMs' moral inference by avoiding the brittleness of direct prompting.

Our methodological contributions are two-fold: first, we present metrics that ensure {\it structural alignment} of LLM- and human-generated word associations to ensure the robustness and reproducibility of our results. Secondly, we introduce a novel moral value propagation algorithm based on a random walk over the {\it global} association network and show that it leads to moral estimates that better correspond to human moral perception than previous work~\cite{Ramezani2024}, which operated on {\it local} sub-graphs.

We identify general patterns of similarity and divergence between LLMs and human participants,\footnote{For the rest of the paper, any comparison between humans and LLMs refers to `English-speaking western cultures' only.} revealing that LLMs and humans align more closely for positive moral values compared to negative ones. Humans show greater emotional diversity and concreteness in their responses, while LLMs are less varied and more abstract. These findings provide critical insights into how LLMs process moral concepts differently from human participants, in the context of western Anglo-centric cultural norms.

In summary, our contributions are as follows:
\begin{itemize}
    \item We are the first to explore moral alignment between humans and LLMs through the lens of the mental lexicon, offering a novel approach to understanding moral alignment.
    \item A framework to effectively extract multidimensional moral values from  global word association networks, allowing for fine-grained evaluation.
    \item A detailed comparison of human and LLM associations, including explanations for divergences along certain dimensions (e.g., fairness and sanctity), in terms of differences in graph structures and varying levels of concreteness and emotionality of generated associations.
\end{itemize}

\section{Background}
\label{sec:background}

\paragraph{Moral Foundation Theory} {(MFT; \citet{graham2013}) is a widely-used framework that attempts to explain human morality through five fundamental and universal dimensions: {\it Care, Fairness, Loyalty, Authority, and Sanctity}. Each dimension is characterized on a scale from vice (-1) to virtue (+1). The Moral Foundations Dictionary~\cite{Frimer2017} which assigns English words along this scale, for each dimension and has been widely used to assess morality in written text. While the original dictionary was expert-created, follow-up work crowd-sourced the extended MFD (eMFD; \citet{Hopp2021}) resulting in a much larger and more diverse set of words associated with moral dimensions. Recent work has re-visited the MFT and proposed to split the {\it fairness} 
dimension into \textit{equality} and \textit{proportionality} to better capture distinct justice motives~\cite{atari2023morality}. We acknowledge that the exact definition of moral foundations are under active research, however, will base our work on the original MFT to directly compare with relevant related work, and to be able to draw on its linguistic resources (MFD and eMFD) to support our study.}

\paragraph{Mental lexicon for moral inference} The {Mental Lexicon} refers to the mental representations and connections of word meanings that support understanding and reasoning~\cite{HField+1981}. It is often conceptualized as a semantic network, where words are represented as nodes and weighted edges reflect their degree of connectivity~\cite{lowe1997,de_deyne2016}. The {\bf Word Association Test} can reveal mental connections by exposing participants to cues (e.g., {\it volunteer}) and asking them for the first words that spring to their mind (e.g., {\it help}, {\it kind} or {\it care}). The obtained results are turned into a {\bf word association graph} with cues and responses as nodes, and edge weights indicating the number of participants who produced a cue-response pair. Prior work has shown that such networks capture basic commonsense knowledge \cite{liu-etal-2021-commonsense,liu-etal-2022-wax} and complex semantics more reliably than direct text-based measures~\cite{de2020cross,de2021visual}, including moral inference~\cite{Ramezani2024}.

\begin{figure*}[ht]
    \centering
    \includegraphics[width=0.95\linewidth]{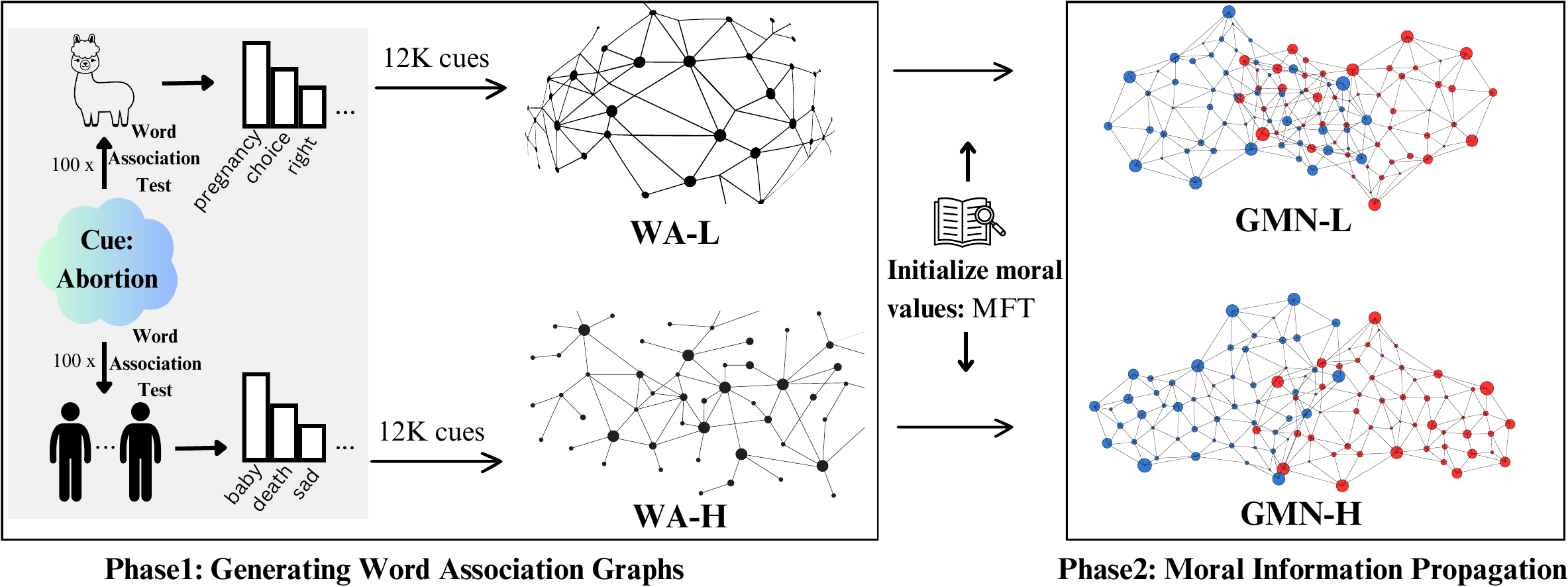}
    \caption{Overview of our two-phase framework: (1) Collecting word association graphs from humans (\wahuman) and Llama (\wallm); (2) Propagating moral information through the word association graphs to obtain two global moral networks  (\wahuman $\rightarrow$ \swow; \wallm $\rightarrow$ \llama), where red and blue nodes indicate words with negative and positive inferred moral scores, respectively.}
    \label{fig: workflow}
    
\end{figure*}

\paragraph{Computational investigations of moral inference}
Moral Association Graphs (MAG) are cognitively motivated models of human moral inference~\cite{Ramezani2024}. Based on human-generated word association networks, the extract local undirected graphs for a given cue word, where nodes are responses and edges are weighted by co-occurrences. Selected responses are seeded with ground truth moral values which are propagated through the local network until convergence. MAG has been shown to be able to predict human moral values, however, MAG operates on {\it local} graphs centered around a single cue which prevents the model to make more complex, long distance interactions. We extend this idea to a {\it global} graph propagation framework where we propagate multi-dimensional moral associations corresponding to the five dimensions of MFT. 

Recent research has applied the word association test to LLMs and investigated similarities and differences to human-generated data sets. \citet{abramski2024} found substantial overlap of node-pairs in the association graphs, but LLMs generated significantly less diverse responses compared to humans, prompting us to explicitly assess response diversity in our experiments. \citet{ramezani-2023-knowledge} demonstrated that LLMs can capture moral norms when prompted directly. However, it remains unclear whether these elicited moral norms reflect a deeper conceptual organization within LLMs regarding morality, or if they are primarily superficial patterns learned from training data that do not necessarily indicate such organization.

\citet{ji2024} applied the widely-used Moral Foundations Questionnaire~\cite{Graham2009} to LLMs, comparing LLM and human responses. They found that LLMs exhibit a superficial understanding of morality, predominantly characterized by phrases they have been exposed to during training, which questions the reliability of their answers\footnote{For example, presenting LLMs with moral statements such as, 'One of the worst things a person could do is hurt a defenseless animal,' followed by a prompt like 'Do you agree with this statement? A. Agree B. Disagree,' encourages LLMs to align with socially accepted norms.} . 
Given their extensive human training data, LLMs are biased towards responses that are widely reported~\cite{anagnostidis2024,Scherrer2023}. Additionally, enforcing a binary response (agree/disagree) prohibits to assess a more nuanced moral reasoning. In contrast, our work probes for moral values indirectly by eliciting conceptual associations from LLMs -- a method that has been shown effective to simulate human moral reasoning~\cite{Ramezani2024}. By reducing the influence of explicit prompting for moral values, our approach minimizes contextual impact.

\section{Framework Overview}
We aim to (1) capture moral values encoded in LLM representations and (2) compare them with human values. We do so in a 3-step framework as shown in Figure \ref{fig: workflow}: First, we obtain spontaneous responses for the same set of 12,000 cues from both humans (using an existing data set from \citet{deDeyne2019}) and LLMs (by prompting with the same set of cues and instructions; Section~\ref{sec:was}). Based on this, we construct a word association graph from human data and another from LLM data. Second, we initialize a `morality score' for selected concepts from a ground truth dataset based on MFT, and propagate this information through the graphs, resulting in two Global Moral Networks (\gmsn, Section~\ref{sec:gmsn}). This \gmsn enables a comparative analysis of moral alignment between humans and the LLM~(Section~\ref{sec:evaluation}).

\subsection{Model and External Datasets} 
\label{sec: dataset}
\paragraph{Model} We used Llama-3.1-8B-Instruct (henceforth Llama) in all our experiments, a state-of-the-art LLM trained over 15 trillion token and including RLHF optimisation~\cite{huang2024}. It was selected due to its performance, accessibility, and good trade-off between computational efficiency and scalability~\cite{dubey2024llama, guo2024}. 

\paragraph{Human Word Associations} We used the English {\it Small World of Words} data set~\cite{deDeyne2019}, which comprises responses from about 90k native English-speaking participants for over 12k cues. We refer to this data set as \wahuman (Word Associations - Human). Each cue was presented to 100 participants, and each participant produced up to three responses, resulting in a broad and varied set of responses. {Participants are primarily English  speakers from the U.S.\ (50\%), as well as the U.K., Canada, and Australia.}

\paragraph{Moral Foundations Dictionary 2.0} (MFD,~\citet{Frimer2017}), which contains 2041 words, assigns selected words to one or more of the five dimensions of the MFT (Section~\ref{sec:background}). Each word is assigned a moral score of 1 if it relates to the dimension's virtue, -1 if it aligns with its vice, and 0 if it is unrelated, leading to a hard assignment of words to moral dimensions. We use the MFD to identify moral seed words in the word association graphs, using the intersection of MFD and {12K cues in word association graphs}, resulting in 626 moral seed words.

\paragraph{Extended Moral Foundations Dictionary} (eMFD; \citet{Hopp2021}) is a crowdsourced extension of MFD. It provides soft associations of English words with one or more of the five moral dimensions, assigning a value between -1 (vice) and 1 (virtue). Following~\citet{Ramezani2024}, we use the eMFD for evaluation. For this, we compare the moral values from eMFD against those predicted by our method for the words found in the intersection of eMFD and our cue word set. This intersection comprises 2,186 words (out of eMFD's 3,270 total words) that are present in our cue set and are used for the correlation comparison (Section ~\ref{ssec:propagation_exp_set_up}).

\section{Eliciting Word Associations from LLMs}
\label{sec:was}

Starting from human word association data set by \citet{deDeyne2019} (henceforth, \wahuman). Then we prompt Llama to obtain a comparable set of LLM-generated word associations which we also transfer into a separate graph (\wallm).

\subsection{Methods}
We prompted Llama to elicit associations with the 12k cues underlying \wahuman. LLM responses are known to be unstable with respect to changes in prompts, and changes in temperature. To address the former, we employ the exact same instructions as used in the \wahuman data collections (full prompt in Appendix~\ref{sec:appendix}) requesting Llama to generate up to three responses per cue, repeating this process 100 times for each cue word, this effectively provides a Monte-Carlo approximation of the probability distribution of word associations.  To ensure validity of our results, we define two criteria for LLM-generated associations: like large-scale human associations, the overall patterns must be {\it robust} and not change significantly should the data be re-collected; in addition, responses should resemble the {\it variability} (or diversity) observed in human associations. We tune Llama's temperature for these objectives.

\paragraph{Temperature tuning} We measure {\it variability} as the total number of distinct word types in Llama's responses over given set of cues. {\it Robustness} is calculated by randomly splitting the responses for each cue in \wallm into two halves and computing the relative word association strength of each response for a given cue in each half.\footnote{The relative word association strength of a response is calculated as \( \text{Strength}_{i} = \frac{f_i}{N} \), where \( f_i \) is the number of times a response \( i \) appears in the cue, and \( N \) is the total number of responses. This measures how strongly a particular response is associated with the cue.} The reliability for a given cue is calculated by Spearman-Brown split-half reliability $ r_{\text{total}} = \frac{2r_{\text{half}}}{1 + r_{\text{half}}}$, where \( r_{\text{half}} \) represents the correlation between association strengths in the two halves~\cite{walker2006,charter1996}. We average \( r_{\text{total}} \) over all selected cues.

\paragraph{Evaluating \wallm}

We evaluate the overlap of responses between \wallm and \wahuman.\footnote{In Appendix \ref{sec:Reliability Test}, we also show a comparison between \wahuman and \wallm in terms of reliability.} We compute precision at $k$ of \wallm responses in the human-produced association set for the same cue with varying $k$. We also report average {correlation} of association strength in \wahuman and \wallm per cue.\footnote{$\frac{1}{n} \sum_{i=1}^{n} \text{cor}(WS_H(i), WS_L(i))$ where \( i \) is a cue word, \( WS_H(i) \) and \( WS_L(i) \) are the human and LLM word association strengths, respectively, for the intersection of responses for cue $i$ in \wahuman and \wallm.} We include a baseline Word2Vec model which associates each cue with the $k$ nearest neighbors in an embedding space based on Google News 300-dimensional embeddings~\cite{mikolov2013}.

\begin{figure}[t]
    %\centering
    \includegraphics[width=0.4\textwidth]{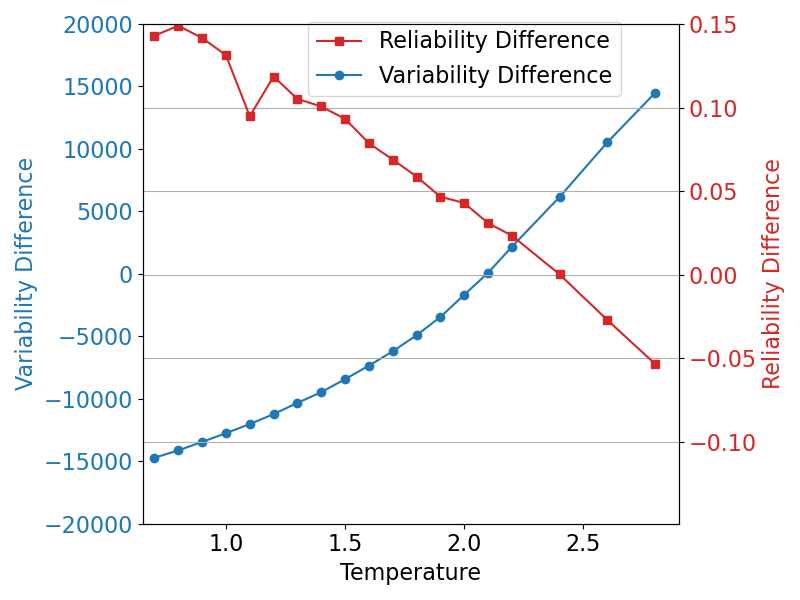}
    \caption{Effect of temperature on differences in variability (blue) and reliability (red) between \wallm and \wahuman (0 is best).}
    \label{fig: varying temperture}
\end{figure}

\subsection{Results}
We tune the temperature based on a random subset of 400 cues. Results in Figure~\ref{fig: varying temperture} show that as the temperature increases, Llama produces more varied responses leading to an increase in diversity and decrease in robustness, both of which approach human values. We generate the full \wallm with the identified optimal temperature of 2.1.

\begin{figure}[t]
    \centering
    \includegraphics[width=0.9\linewidth]{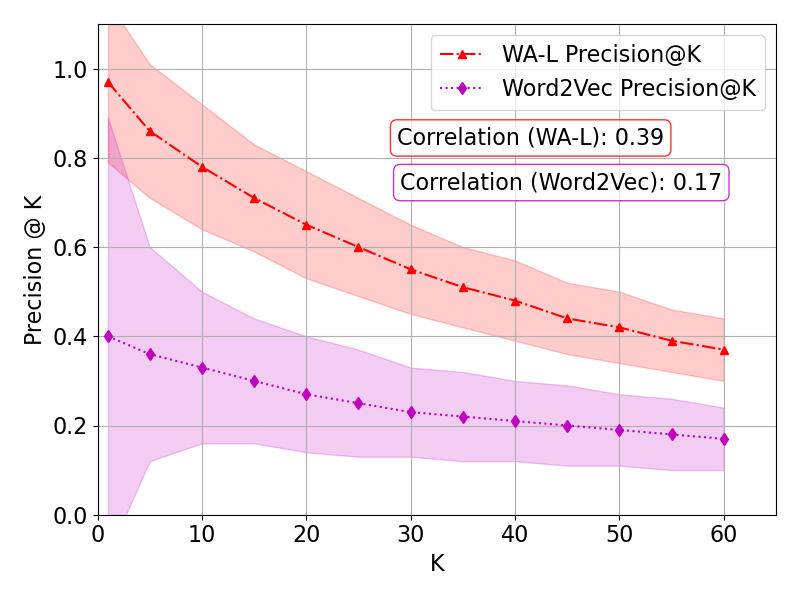}\hfill
    \caption{Precision@K for \wallm, and Word2Vec Associations relative to \wahuman. Shaded regions show standard deviation over 50 runs. Correlation scores are noted.}
    \label{fig:precision@k}
\end{figure}

For the evaluation of our final \wallm we select 279 cues from the MFD, ensuring equal representation of verbs, adjectives, and nouns.\footnote{The smallest POS class are adjectives with only 93 instances. POS tags were obtained with spaCy.} 
We focus on cues from the MFD to specifically assess agreement on this domain of interest. Figure~\ref{fig:precision@k} shows that \wallm almost perfectly agrees with the most frequent response for a moral cue ($k=1$), with the precision slowly decreasing just below 80\% agreement for the top 10 cues. Precision declines further as \( k \) increases, reflecting the divergence between Llama's broader set of moral associations and \wahuman responses. The Word2Vec baseline leads to noticeably worse precision, particularly for small~$k$. Appendix \ref{sec:Graph Statistic} provides statistics for  \wahuman and \wallm.

\section{Global Moral Networks}
\label{sec:gmsn}
\wahuman and \wallm reflect how words are interconnected in human and LLM representations, but do not inherently encode moral scores. {We now propagate moral values through the \wahuman and \wallm networks to predict moral associations {scores} of concepts with each of the five MFT dimensions.} We propagate moral information separately through each network obtaining two {\bf Global Moral Networks} (\gmsn): \swow (propagated from \wahuman) and \llama (propagated from \wallm).

\subsection{Moral Information Propagation}

Our word association graph {$G \in \{ \wahuman, \wallm \}$} consists of $|n|$ nodes and $|\epsilon|$ edges, and we aim to assign a five-dimensional moral value vector to each node $c_i$ to obtain a \gmsn.
We represent the moral values in a matrix $F \in \mathbb{R}^{|n| \times 5}$, where each row represents a cue word $c_i$ from $G$, and columns are the five moral dimensions. Initially, all elements in $F_0$ are set to zero.
We then initialize $F_0$ with moral values by assigning each $c_i \in$ MFD its five associated moral values $\in [-1, 1, 0]$ for vice, virtue and no association, respectively. 
This moral information is spread iteratively to the entire graph using a random walk \cite{Zhou2003, du2019}: 

\begin{equation*}
    F_{t+1} = \alpha S F_t + (1 - \alpha) F_0,
\end{equation*}
where 
\begin{equation*}
    S = D^{-\frac{1}{2}} W D^{-\frac{1}{2}} \in \mathbb{R}^{|n| \times |n|}
\end{equation*}

$W$ is the adjacency matrix of the word association graph $G$, and the diagonal matrix $D$ contains the sum of the corresponding row values in $W$.
$\alpha \in (0, 1) $ is a hyperparameter that controls the extent of propagation in the graph, with smaller values pulling the local connections closer to the initial matrix $F_0$. This process assigns a 5-dimensional moral value to all words in the \gmsn.\footnote{Practically, we use the closed-form solution proposed in~\citet{Zhou2003} $F^* = (I - \alpha S)^{-1} F_0$, where $I$ is the identity matrix.}

\subsubsection{Experimental Setup}
\label{ssec:propagation_exp_set_up}
\paragraph{Optimizing alpha} We use the portion of the eMFD which is {\it not} used in evaluation, obtaining 277 words with eMFD labels and optimize the correlation between predicted and eMFD moral values.\footnote{In the MAG experiment, the total number of distinct words from each dimension is 1,909. This set of 1,909 words is used for later evaluation, while the 277 (2,186 - 1,909) words are used for tuning alpha.} We find that \swow requires a smaller $\alpha{=}0.75$ for optimal performance, while \llama performs best at $\alpha{=}0.9$ (detailed in the Appendix \ref{sec:appendix2}). A higher $\alpha$ promotes stronger propagation, suggesting \llama might be less efficient at transmitting information. This is supported by graph statistics: the human graph has a smaller diameter\footnote{The length of the shortest path between the most distant nodes.} (3 vs. 4), higher density (0.013 vs. 0.007), and higher connectivity (114 vs. 77), indicating that information can diffuse through it more easily~\cite{taxidou2014,centola2010}, hence needing a lower $\alpha$ for effective propagation.

From a robustness perspective, results in Appendix~\ref{sec:appendix2} suggest limited sensitivity of the propagation algorithm to alpha, indicating stability up to a threshold where performance decreases rapidly.

\paragraph{Evaluation} Following the propagation process, we obtain moral scores across five dimensions for each of the 12,000 cues in both \llama and \swow. To assess the alignment of these moral scores with MFT, we measure the Spearman correlation between our propagated scores and human-annotated moral scores in the eMFD. To measure the generalizability of propagation on new concepts, we subtract the seed values from all nodes which were part of the MFD initialization. We compare against the state-of-the-art model MAG \cite{Ramezani2024}, which has been shown to outperform  Word2Vec and GPT-3.5 on the same task.
\subsection{Results: Concept Morality Prediction}
\begin{table}[t]
    \centering
    \setlength{\tabcolsep}{3pt}
    \begin{small}
    \begin{tabular}{lccc}
        \toprule
        \textbf{Moral Dimension} & \textbf{MAG} & \textbf{\swow} & \textbf{\llama} \\\midrule
        {Care (n = 1895)} & 0.29 & \textbf{0.47} & 0.46 \\ 
        {Sanctity (n = 1893)} & 0.25 & 0.39 & \textbf{0.44} \\ 
        {Fairness (n = 1514)} & 0.23 & 0.29 & \textbf{0.32} \\ 
        {Authority (n = 1737)} & 0.21 & 0.19 & \textbf{0.25} \\ 
        {Loyalty (n = 1714)} & \textbf{0.30} & 0.26 & \textbf{0.30} \\ \midrule
        {All  (n = 8753)} & 0.20 & 0.28 & \textbf{0.29} \\\bottomrule
    \end{tabular}
    \end{small}
    \caption{Correlation of predicted moral values against the eMFD. MAG and \swow are run on the same underlying graph (\wahuman) while \llama ran on \wallm. $n$ indicates the number of concepts per dimension, and overall. All correlations are statistically significant (p $\leq 0.01$). }
    \label{tab:moral_foundations}
\end{table}

\begin{table*}[htbp]
    \centering
    \begin{small}
    \begin{tabular}{c c|c c|cc}
        \toprule
        \multicolumn{2}{c|}{\textbf{Negative}}  & \multicolumn{2}{c|}{\textbf{Positive}} & \multicolumn{2}{c}{\textbf{Different}} \\ 
        \textbf{{\swow}} &  \textbf{{\llama}} & \textbf{{\swow}} &  \textbf{{\llama}} & {\textbf{\llama}$\uparrow$ \textbf{\swow}$\downarrow$} & {\textbf{\llama}$\downarrow$ \textbf{\swow}$\uparrow$} \\\midrule
        \textbf{disgusting} &  betrayal & \underline{\textbf{church}} &  \underline{\textbf{church}} & \hspace{1.45 em}\text{abortion}\hspace{1.4 em} & jail\\
        traitor &  \underline{prejudice} & \textbf{religion} &  \underline{kindness} & \hspace{1em}\text{immigrant}\hspace{1em}  & air\\
        \underline{vomit} &  cheating & God &  \textbf{religion} & \hspace{1.1em}\text{politician}\hspace{1.3em} & plastic\\
        hurt &  \textbf{disgusting} & \textbf{priest} &  \textbf{priest} & \hspace{1.2em}\text{capitalist}\hspace{1.3em} & Soviet\\
        dirty &  \underline{discrimination} & {holy} &  prayer & \hspace{0.5em}\text{homosexual} & bees\hspace{0.8em}\\
        pain &  dishonest & religious &  bible & \hspace{0.6em}\text{commercial}\hspace{0.8em} & snob \\\bottomrule
    \end{tabular}
    \end{small}
    \caption{Comparison of top negative, positive, and most different concepts between \llama and \swow. Common concepts are \textbf{bolded}. Responses from the two methods for the \underline{underlined} concepts are given in Table~\ref{tab:negative_response_main}. The Difference block shows concepts rated significantly more positive by the \llama compared to \swow (left) and vice versa (right). Moral values for these concepts, along with other top 10 negative and positive moral concepts, are provided in Appendix \ref{sec:Ranking Values}. }
    \label{tab:comparison_table}
\end{table*}
\begin{table}[t]
    \centering
    \small
    
    % \resizebox{\textwidth}{!}{%
    \begin{tabular}{c|p{1.8cm}p{2.4cm}}
        \toprule
        \textbf{Concept} & \multicolumn{2}{c}{\textbf{Top Unique Responses}} \\
        & \textbf{\swow} & \textbf{\llama} \\ \midrule
        \multirow{3}{*}{\textbf{prejudice} }
        & pride, black, race, racist & stereotypes, biases, stereotyping, \newline bigoted \\\midrule

        \multirow{3}{*}{\textbf{discrimination}}
        & race, racist, sexism, \newline gender & stereotypes, stereotyping, equality, prejudices \\\midrule

        \multirow{3}{*}{\textbf{vomit}}
        & gross, spew, smell, green & stomachache, queasy, hangover, poisoning \\\midrule\midrule 
        
        \multirow{2}{*}{\textbf{kind} }
        & type, sort, happy, person & nurturing, soft, charitable, warmth \\\midrule

        \multirow{3}{*}{\textbf{church}}
        & catholic, synagogue, stone, school & altar, minister, \newline baptism, service \\\bottomrule

    \end{tabular}%
    % }
    \caption{Comparison of the top four unique responses between \swow and \llama for highly negative (top) and positive (bottom) moral concepts.}
    \label{tab:negative_response_main}
\end{table}

Table~\ref{tab:moral_foundations} presents our experimental results. 
Overall, our propagated moral scores demonstrate higher correlation with human judgments than MAG. This stronger positive correlation highlights the effectiveness of global graph propagation, in contrast to MAG's local, cue-specific graphs (see Section~\ref{sec:background}). We attribute this improved performance to the importance of multi-hop propagation over longer distances in the network. For instance, our model effectively captures the association between ``mother''  and ``life''  through intermediate concepts such as ``birth''. This demonstrates how our model captures the nuanced relationships between seemingly different concepts, reflecting a more comprehensive understanding of moral concepts.\footnote{Figure \ref{fig:p1} shows an example of the propagation process.}

Our two association graphs, \llama and \swow exhibit comparable overall correlation with the eMFD, but differ across individual dimensions, with the largest differences observed for {\it sanctity} and {\it authority}.\footnote{We assume that \llama exceeds \swow because both the eMFD and LLMs like Llama are heavily based on text-based knowledge while human associations reflect a broader range for modality and experience, something we dig in to in the following sections.} This is interesting, as it indicates where humans and LLMs diverge, however, it does not explain why these differences exist. We next qualitatively analyze these differences and uncover systematic underlying factors.

\section{Moral Alignment between Humans and LLMs}
\label{sec:evaluation}
After evaluating the reliability and robustness of our framework, we proceed to assess moral alignment between \swow and \llama using propagated values derived from our approach.

\subsection{Cross-Dimensional Analysis}
We start our analysis by investigating the moral alignment between \swow and the \llama on the overall moral perception on concepts. We calculate each concept's \textbf{overall morality} by summing its moral scores across the five dimensions for both positive (virtues) and negative (vices), then rank the concepts accordingly. From these ranked lists, we select representative samples and analyze their responses  within each moral dimension to observe the patterns of \swow and \llama. Lastly, we build local subgraphs for the top 50 negative words in each dimension to understand propagation efficiency using density and weighted average edge.

\begin{table*}[htbp]
    \centering
    \begin{small}
    \begin{tabular}{l|cl|cl|cl|cl|cl|cl}
        \toprule
        & \multicolumn{2}{c|}{\textbf{Care}} & \multicolumn{2}{c|}{\textbf{Fairness}} & \multicolumn{2}{c|}{\textbf{Loyalty}} & \multicolumn{2}{c|}{\textbf{Authority}} & \multicolumn{2}{c|}{\textbf{Sanctity}} & \multicolumn{2}{c}{\textbf{All}} \\ 
        & \textbf{H} & \textbf{L}  & \textbf{H} & \textbf{L}  & \textbf{H} & \textbf{L}  & \textbf{H} & \textbf{L} &  \textbf{H} & \textbf{L} &  \textbf{H} & \textbf{L} \\ \midrule
        \# Moral Concepts & \multicolumn{2}{c|}{70} & \multicolumn{2}{c|}{68} & \multicolumn{2}{c|}{60} & \multicolumn{2}{c|}{65} & \multicolumn{2}{c|}{70} & \multicolumn{2}{c}{6941}\\ \midrule
        Emotional responses (\%) & \textbf{72} & 61* & \textbf{67} & 54*  & \textbf{69} & 54*  & \textbf{67} & 59*  & \textbf{69} & 58*  & \textbf{66} & 55*  \\
        Emotional intensity & 4.24 & 4.1  & 3.71 & 3.77  & {3.8} & 3.82  & 3.78 & \textbf{4.10}* & \textbf{3.81} & 3.60* & \textbf{3.30} & 3.17* \\ \midrule
        Concrete responses (\%) & \textbf{35} & 24* & \textbf{24} & 12* & \textbf{24} & 12* & \textbf{29} & 16* & \textbf{40} & 33* & \textbf{42} & 36* \\ 
        Concreteness score & \textbf{3} & 2.7* & \textbf{2.6} & 2.2* & \textbf{2.5} & 2.3* & \textbf{2.7} & 2.5* & \textbf{3.2} & 3* & \textbf{3.1} & 2.9* \\
        \bottomrule
    \end{tabular}%
    \end{small}
    \caption{Average proportion of emotional responses and intensity (top), and concrete responses and concreteness scores (bottom) in the top 50 negative cues from \swow(H) and \llama (L)-generated responses. The concepts are associated with moral dimensions identified by both humans and the LLM. The comparison size of moral concepts is the union of H and L from their respective top words.  * indicates statistically significant differences (t-test; $p<0.05$). Significantly higher scores are bolded.}
    \label{tab:comparison_analysis}
\end{table*}

\paragraph{Results}
Table \ref{tab:comparison_table} presents the top positive and negative moral concepts for \swow and \llama. \swow's top negative concepts often relate to physically or emotionally charged words in the sanctity dimension (e.g., ``disgusting'', ``gross''), whereas \llama focuses predominantly on social vices in the fairness dimension (e.g., ``betrayal'', ``racism''). Despite these differences, both \swow and \llama significantly overlap in top positive concepts which refer to virtuous or religious concepts. In several instances \swow and \llama moral scores diverged in polarity such as ``abortion'', ``capitalist'' (humans more negative than Llama) or ``plastic'' (humans more positive than Llama). 

\emph{Why do the top negative concepts diverge between \llama and \swow?} We inspected the local graph topology around the most negative \textit{abstract} \llama concepts (like ``prejudice'', or ``discrimination'') and find a dense network\footnote{These words often appear as top responses to each other.} of abstract (thematic or causal) connections among these concepts. Associations for ``prejudice'' and ``discrimination'' are shown in Table~\ref{tab:negative_response_main}; more examples in Appendix~\ref{sec:negative_response_analysis}). These associations are reflective of systemic discussions captured in the model's training data \cite{fish2020racism, baldwin2017culture, dai2024, zheng2023, tjuatja2024, dillion2023ai}. In contrast, \swow associations for the same concepts are more varied, often influenced by individual sensory experiences and cultural context \cite{kostova2008word, son2014understanding, shin2018happiness}. For example, the concept ``prejudice'' is frequently associated with culturally specific concepts like ``race'' or ``black'' resulting in divergent semantic networks compared to \llama's statistically driven associations (e.g., ``bigotry''). When considering negative \textit{physical or emotional} concepts like ``vomit'' \swow consistently involves synonymous concepts, indicative of direct sensory or emotional experiences (see ``vomit'' in Table~\ref{tab:negative_response_main} and more examples in Appendix~\ref{sec:negative_response_analysis}). In contrast, \llama still maintains a focus on causal relations. This discrepancy highlights a systematic qualitative difference between representations based on statistical word co-occurrence patterns~\cite{kang2023} and the rich associations observed in humans reflecting their rich physical and emotional experience~\cite{ji2024}. This difference clearly persists in associations, although in the dialogue tasks that LLMs increasingly approach human capabilities. 

In positive moral concepts, we observe that responses from both \swow and \llama to virtue-related words often display synonymy or antonymy, while religion-related concepts exhibit various types of meronymy (Table~\ref{tab:negative_response_main} bottom, and Appendix \ref{sec:positive response analysis}). Llama is predominantly trained on training data from Western cultures, where
religious concepts have a strong, positive historical presence despite the declining influence of religion in many Western societies \cite{topkev2024, halman1994religion}. This cultural frameworks naturally lead to overlap in positive moral concepts between humans and Llama.\footnote{An interesting direction for future work is to replicate these experiments with LLMs trained on corpora from secular societies (e.g., China) or societies dominated by religions other than Christianity.}

The quantitative analysis of subgraphs across dimensions reveals several important findings (statistical details are provided in Appendix \ref{sec: subgraph}). First, the statistics suggest that moral words associated with the fairness and sanctity dimensions in \llama exhibit stronger propagation efficiency (higher weighted average edge centrality) and are more densely connected in the fairness dimension, leading to significant advantages in spreading moral information ~\cite{taxidou2014,centola2010}.\footnote{Even though the difference may not be significantly larger than in other dimensions, these advantages could accumulate across multiple metrics.} Moreover, \llama demonstrates notably stronger connections within other abstract dimensions such as loyalty and authority, with weighted degree centrality being two times higher than \swow, while the magnitude is similar in the care and sanctity dimensions. Finally, both propagation efficiency and density decrease significantly when pruning the graph to retain only the top moral words for both \swow and \llama, suggesting that morally significant concepts across dimensions are highly interconnected and exhibit stronger propagation efficiency compared to less morally related concepts.

\subsection{Human moral associations are more emotional and concrete}
We identified systematic qualitative differences in the associations with morally negatively connotated cues (vices). Specifically, Llama associations with morally loaded words are more sterile with less emotion and a higher level of abstractness.

\paragraph{Method} We analyze {\bf emotionality} in responses to the top 50 morally significant concepts across five moral dimensions. We obtain an emotion score for each response using the {\it arousal} norms from the VAD-norms~\citep{warriner2013norms}, a human-labeled emotion lexicon of over 13k English words.\footnote{In this lexicon, a score close to 1 suggests that the concept tends to evoke a relaxed, bored, or sleepy emotional state, indicating a low arousal response, whereas a score near 8 signifies that the concept tends to be associated with feelings of excitement, happiness, or high arousal.} We quantify the degree of emotions reflected in responses per cue using (a) the proportion of emotional responses among all responses and (b) their average emotional intensity. A response is considered emotional if it is in the emotion lexicon. Emotional response intensity per concept was calculated by multiplying the emotional intensity of responses by their word association strength, then averaging these values for each moral dimension. The \textbf{concreteness} of responses was assessed using the \citet{brysbaert2014concreteness} concreteness lexicon.\footnote{Highly concrete concepts (a score within the range of 4 to 5) are defined as those that can be directly experienced through the senses, such as objects, actions, or sensations that are easily demonstrable (e.g., ``sweet'' as experienced by tasting sugar). In contrast, abstract concepts (a score within the range of 1 to 3) refer to those that cannot be directly experienced through the senses or actions, such as``prejudice'' in the context of fairness or ``leader'' in the context of authority.} The lexicon contains 37,058 concepts, concepts with a score above 3.5 were considered concrete. The same set of concepts and comparison size from the emotion analysis was used to maintain consistency. We calculated concept-level  concreteness analogously to emotional intensity.\footnote{85\% and 98\% of the top 50 cue words are found in the emotion and concreteness dictionaries, respectively. Nevertheless, even if a cue word is not in these dictionaries, we can still calculate its corresponding scores because these scores are derived from its associated responses, a significant portion of which typically appears in the dictionaries.} 

\paragraph{Results} 
Table \ref{tab:comparison_analysis} presents a detailed comparison of the results. \swow exhibits a consistently higher proportion of unique emotional responses across all dimensions, indicating that it generally provides more diverse emotional responses on average. Additionally, \swow shows higher emotional intensity for sanctity-related dimensions, suggesting that concepts associated with emotional or physical states are more likely to elicit a strong emotional response from \swow compared to \mbox{\llama}. Conversely, for abstract concepts, which are often represented in the fairness, loyalty, and authority dimensions, \swow are less likely to show highly intense emotional responses compared to \llama.  Furthermore, when examining all top negative words, which include a substantial number of morally less significant concepts, we observe a lower average emotional intensity compared to the top 50 negative moral values across dimensions for both \swow and \llama. This suggests a positive correlation between moral significance and emotional intensity in responses.

In the concreteness experiment, \swow tends to produce more concrete responses, whereas \llama's responses are generally more abstract. As observed in Appendix~\ref{sec:negative_response_analysis}, \swow frequently connects cues to real-life or physical experiences~\cite{kostova2008word, son2014understanding, shin2018happiness}. In contrast, Llama relies on abstract associations derived from textual data~\cite{ji2024,Scherrer2023}. This reliance on statistical, text-based associations limits its ability to replicate the sensory-driven responses typical of humans, which dominate moral word associations. Consequently, Llama's responses exhibit lower concreteness scores and less variation overall~\cite{dillion2023ai, Santurkar2023refelectingopinions}

\section{Conclusions}
We presented a framework for a detailed comparison of moral associations between English-speaking, western populations and LLMs, introducing a method to elicit word associations from LLMs that ensures structural similarity to human responses. Our findings demonstrate that moral perspectives can be uncovered through these associations without direct moral prompting, with Llama's moral associations broadly aligning with human performance. The use of a global network approach enabled us to capture nuanced relationships between moral concepts. A key finding is the considerable alignment in top positive moral concepts, likely reflecting shared cultural frameworks. This alignment suggests that LLM representations do reflect aspects of human moral conceptualizations. If such alignment can be consistently achieved, it could further ensure the safety of deploying trustworthy AI when navigating morally-tinged scenarios. However, we also observed notable divergences, particularly among top negative moral concepts. Humans exhibit sensory and experience-driven associations, which are more grounded and emotional. In contrast, LLMs tend towards more abstract concepts with lower emotional intensity, particularly for physical or mental states. These divergences highlight the potential risks of current LLMs operating with a misaligned `moral map'. An LLM lacking the experiential, affective grounding for negative concepts might misjudge the severity of harm or respond inappropriately in critical situations, even if capable of superficially correct answers to direct queries. 

Overall, while LLMs mirror moral associations in English western cultures in many respects, internal processing differences can lead to significant divergences. Our framework provides a valuable tool for identifying these areas. Future work can apply our framework across a wider range of models and to different cultures. Crucially, further research could explicitly link these associative alignments and misalignments to observable LLM behaviors in ethically relevant tasks, thereby deepening our understanding of the critical question of how foundational conceptual structures translate to practical human-LLM alignment.

\section{Limitations}
\paragraph{LLM Selection} As our main focus is to explore the feasibility of automatically generating reliable large-scale word associations and comparing morality alignment, we selected the recent representative Llama-3.1-8B model given its balance of performance and size in various NLP tasks. We acknowledge that different models might exhibit different behaviors. {However, our study is designed as a proof of concept for a framework that is adaptable to different language models. The proposed three-step framework---comprising word association generation, graph-based propagation of moral values, and comparative analysis---is not reliant on any specific LLM. Thus, the methods and insights developed in this study can be applied to other models. While variations in outputs may arise, these differences reflect the inherent diversity of the models being evaluated rather than any limitation of the framework itself.} We leave the exploration of more large language models with varying sizes and types as future work.

\paragraph{Cultural specificity} Moral values vary across cultures~\cite{atari2023morality} and our study only covers Western, English-speaking cultures because both the human participants that generated \wahuman as well as the training data for Llama3.1-8b predominantly originate from this culture. We emphasize this focus in our paper. However, human word association data sets exist for other countries, too~\cite{deDeyne2019} and LLMs are currently developed in and adapted to many languages and communities. While we make no universal claims, we believe that our method enables cross-cultural studies in the future.

\paragraph{Concept-Level Alignment} Our study focuses on providing a framework to systematically evaluate the moral alignments between concepts in humans and LLMs. This approach is not directly applicable to assess morality alignment in broader contexts, such as sentences or documents, where the overall morality is complex to predict. However, the propagated moral scores for large-scale concepts can serve as basic, word-level scores,  supporting future work on contextual moral inference. 

\section*{Acknowledgements}
We thank Aida Ramezani and Yang Xu for sharing their code and data. LF is supported by the ARC Discovery Early Career Research Award (Grant No. DE230100761). SDD is supported by the ARC Discovery Project Research Award (Grant No. DP240101873).

\bibliography{custom}

\begin{thebibliography}{52}
\providecommand{\natexlab}[1]{#1}

\bibitem[{Abdulhai et~al.(2023)Abdulhai, Serapio-Garcia, Crepy, Valter, Canny, and Jaques}]{abdulhai2023}
Marwa Abdulhai, Gregory Serapio-Garcia, Clément Crepy, Daria Valter, John Canny, and Natasha Jaques. 2023.
\newblock \href {https://arxiv.org/abs/2310.15337} {Moral foundations of large language models}.
\newblock \emph{arXiv preprint arXiv:2310.15337}.

\bibitem[{Abramski et~al.(2024)Abramski, Lavorati, Rossetti, and Stella}]{abramski2024}
Katherine Abramski, Clara Lavorati, Giulio Rossetti, and Massimo Stella. 2024.
\newblock \href {https://doi.org/10.3233/FAIA240177} {Llm-generated word association norms}.
\newblock In \emph{HHAI 2024: Hybrid Human AI Systems for the Social Good}, pages 3--12. IOS Press.

\bibitem[{Almeida et~al.(2024)Almeida, Nunes, Engelmann, Wiegmann, and de~Araújo}]{Almeida2024}
Guilherme F. C.~F. Almeida, José~Luiz Nunes, Neele Engelmann, Alex Wiegmann, and Marcelo de~Araújo. 2024.
\newblock \href {https://doi.org/10.1016/j.artint.2024.104145} {Exploring the psychology of llms’ moral and legal reasoning}.
\newblock \emph{Artificial Intelligence}, 333:104145.

\bibitem[{Anagnostidis and Bulian(2024)}]{anagnostidis2024}
Sotiris Anagnostidis and Jannis Bulian. 2024.
\newblock \href {https://doi.org/10.48550/arXiv.2408.11865} {How susceptible are llms to influence in prompts?}
\newblock \emph{arXiv preprint arXiv:2408.11865}.
\newblock Computer Science > Computation and Language (cs.CL).

\bibitem[{Atari et~al.(2023)Atari, Haidt, Graham, Koleva, Stevens, and Dehghani}]{atari2023morality}
M.~Atari, J.~Haidt, J.~Graham, S.~Koleva, S.~T. Stevens, and M.~Dehghani. 2023.
\newblock \href {https://doi.org/10.1037/pspp0000470} {Morality beyond the weird: How the nomological network of morality varies across cultures}.
\newblock \emph{Journal of Personality and Social Psychology}, 125(5):1157--1188.

\bibitem[{Baldwin(2017)}]{baldwin2017culture}
John Baldwin. 2017.
\newblock \href {https://oxfordre.com/communication/view/10.1093/acrefore/9780190228613.001.0001/acrefore-9780190228613-e-164} {Culture, prejudice, racism, and discrimination}.
\newblock \emph{Oxford Research Encyclopedia of Communication}.
\newblock Date of access 5 Oct. 2024.

\bibitem[{Brysbaert et~al.(2014)Brysbaert, Warriner, and Kuperman}]{brysbaert2014concreteness}
Marc Brysbaert, Amy~Beth Warriner, and Victor Kuperman. 2014.
\newblock Concreteness ratings for 40 thousand generally known english word lemmas.
\newblock \emph{Behavior Research Methods}, 46(3):904--911.

\bibitem[{Centola(2010)}]{centola2010}
Damon Centola. 2010.
\newblock \href {https://doi.org/10.1126/science.1185231} {The spread of behavior in an online social network experiment}.
\newblock \emph{Science}, 329:1194--1197.

\bibitem[{Charter(1996)}]{charter1996}
Richard~A. Charter. 1996.
\newblock \href {https://doi.org/10.2466/pms.1996.82.2.401} {Note on the underrepresentation of the split-half reliability formula for unequal standard deviations}.
\newblock \emph{Perceptual and Motor Skills}, 82(2):401--402.

\bibitem[{Clark(1970)}]{clark1970word}
Herbert~H Clark. 1970.
\newblock Word associations and linguistic theory.
\newblock \emph{New horizons in linguistics}, 1:271--286.

\bibitem[{Dai et~al.(2024)Dai, Xu, Xu, Pang, Dong, and Xu}]{dai2024}
Sunhao Dai, Chen Xu, Shicheng Xu, Liang Pang, Zhenhua Dong, and Jun Xu. 2024.
\newblock \href {https://doi.org/10.1145/3637528.3671458} {Bias and unfairness in information retrieval systems: New challenges in the llm era}.
\newblock In \emph{Proceedings of the 30th ACM SIGKDD Conference on Knowledge Discovery and Data Mining (KDD '24)}, page 11 pages, New York, NY, USA. ACM.

\bibitem[{De~Deyne et~al.(2020)De~Deyne, Cabana, Li, Cai, and McKague}]{de2020cross}
Simon De~Deyne, {\'A}lvaro Cabana, Bing Li, Qing Cai, and Meredith McKague. 2020.
\newblock A cross-linguistic study into the contribution of affective connotation in the lexico-semantic representation of concrete and abstract concepts.
\newblock In \emph{CogSci}.

\bibitem[{De~Deyne et~al.(2016)De~Deyne, Kenett, Anaki, and Faust}]{de_deyne2016}
Simon De~Deyne, Yoed~N. Kenett, David Anaki, and Miriam Faust. 2016.
\newblock Large-scale network representations of semantics in the mental lexicon.
\newblock In Michael Ramscar, Matt Jones, Melody Dye, and Ernest Klein, editors, \emph{Big Data in Cognitive Science}, 1st edition, page~7. Psychology Press.

\bibitem[{De~Deyne et~al.(2021)De~Deyne, Navarro, Collell, and Perfors}]{de2021visual}
Simon De~Deyne, Danielle~J Navarro, Guillem Collell, and Andrew Perfors. 2021.
\newblock Visual and affective multimodal models of word meaning in language and mind.
\newblock \emph{Cognitive Science}, 45(1):e12922.

\bibitem[{Deyne et~al.(2019)Deyne, Navarro, Perfors, Brysbaert, and Storms}]{deDeyne2019}
Simon~De Deyne, Danielle~J. Navarro, Amy Perfors, Marc Brysbaert, and Gert Storms. 2019.
\newblock The “small world of words” english word association norms for over 12,000 cue words.
\newblock \emph{Behavior Research Methods}, 51(3):987--1006.

\bibitem[{Dillion et~al.(2023)Dillion, Tandon, Gu, and Gray}]{dillion2023ai}
Danica Dillion, Niket Tandon, Yuling Gu, and Kurt Gray. 2023.
\newblock \href {https://doi.org/10.1016/j.tics.2023.04.008} {Can ai language models replace human participants?}
\newblock \emph{Trends in Cognitive Sciences}, 27(7):597--600.

\bibitem[{Du et~al.(2019)Du, Wu, and Lan}]{du2019}
Yupei Du, Yuanbin Wu, and Man Lan. 2019.
\newblock Exploring human gender stereotypes with word association test.
\newblock In \emph{Proceedings of the 2019 Conference on Empirical Methods in Natural Language Processing and the 9th International Joint Conference on Natural Language Processing (EMNLP-IJCNLP)}, pages 6133--6143, Hong Kong, China. Association for Computational Linguistics.

\bibitem[{Dubey et~al.(2024)Dubey, Jauhri, Pandey, Kadian, Al-Dahle, Letman, Mathur, Schelten, Yang, Fan et~al.}]{dubey2024llama}
Abhimanyu Dubey, Abhinav Jauhri, Abhinav Pandey, Abhishek Kadian, Ahmad Al-Dahle, Aiesha Letman, Akhil Mathur, Alan Schelten, Amy Yang, Angela Fan, et~al. 2024.
\newblock The llama 3 herd of models.
\newblock \emph{arXiv preprint arXiv:2407.21783}.

\bibitem[{Field(1981)}]{HField+1981}
Hartry~H. Field. 1981.
\newblock \href {https://doi.org/doi:10.4159/harvard.9780674594722.c7} {\emph{5. Mental Representation}}, pages 78--114.
\newblock Harvard University Press, Cambridge, MA and London, England.

\bibitem[{Fish and Syed(2020)}]{fish2020racism}
Jillian Fish and Moin Syed. 2020.
\newblock \href {https://www.researchgate.net/profile/Jillian-Fish/publication/339240954_Racism_Discrimination_and_Prejudice/links/5e45a7c5299bf1cdb9284646/Racism-Discrimination-and-Prejudice.pdf} {Racism, discrimination, and prejudice}.
\newblock In \emph{The Encyclopedia of Child and Adolescent Development}, pages 1--12. John Wiley \& Sons, Inc.

\bibitem[{Frimer et~al.(2017)Frimer, Haidt, Graham, Dehghani, and Boghrati}]{Frimer2017}
Jeremy Frimer, Jonathan Haidt, Jesse Graham, Morteza Dehghani, and Reihane Boghrati. 2017.
\newblock Moral foundations dictionaries for linguistic analyses, 2.0.
\newblock Unpublished Manuscript.

\bibitem[{Graham et~al.(2013)Graham, Haidt, Koleva, Motyl, Iyer, Wojcik, and Ditto}]{graham2013}
Jesse Graham, Jonathan Haidt, Sena Koleva, Matt Motyl, Ravi Iyer, Sean~P. Wojcik, and Peter~H. Ditto. 2013.
\newblock Moral foundations theory: The pragmatic validity of moral pluralism.
\newblock In Patricia Devine and Ashby Plant, editors, \emph{Advances in Experimental Social Psychology}, volume~47, pages 55--130. Academic Press.

\bibitem[{Graham et~al.(2009)Graham, Haidt, and Nosek}]{Graham2009}
Jesse Graham, Jonathan Haidt, and Brian~A. Nosek. 2009.
\newblock \href {https://doi.org/10.1037/a0015141} {Liberals and conservatives rely on different sets of moral foundations}.
\newblock \emph{Journal of Personality and Social Psychology}, 96(5):1029--1046.

\bibitem[{Guo et~al.(2024)Guo, Farnan, McLaughlin, and Devereux}]{guo2024}
Rui Guo, Greg Farnan, Niall McLaughlin, and Barry Devereux. 2024.
\newblock \href {https://doi.org/10.18653/v1/2024.bionlp-1.58} {{QUB}-cirdan at {\textquotedblleft}discharge me!{\textquotedblright}: Zero shot discharge letter generation by open-source {LLM}}.
\newblock In \emph{Proceedings of the 23rd Workshop on Biomedical Natural Language Processing}, pages 664--674, Bangkok, Thailand. Association for Computational Linguistics.

\bibitem[{Halman and De~Moor(1994)}]{halman1994religion}
Loek Halman and Ruud De~Moor. 1994.
\newblock Religion, churches and moral values.
\newblock In \emph{The individualizing society}, pages 37--65. Brill.

\bibitem[{Henrich et~al.(2010)Henrich, Heine, and Norenzayan}]{henrich2010weirdest}
Joseph Henrich, Steven~J Heine, and Ara Norenzayan. 2010.
\newblock The weirdest people in the world?
\newblock \emph{Behavioral and brain sciences}, 33(2-3):61--83.

\bibitem[{Hopp et~al.(2021)Hopp, Fisher, Cornell, Huskey, and Weber}]{Hopp2021}
Frederic~R. Hopp, Jacob~T. Fisher, Devin Cornell, Richard Huskey, and René Weber. 2021.
\newblock The extended moral foundations dictionary (emfd): Development and applications of a crowd-sourced approach to extracting moral intuitions from text.
\newblock \emph{Behavior Research Methods}, 53(1):232--246.

\bibitem[{Huang et~al.(2024)Huang, Zheng, Ma, Qin, Lv, Chen, Luo, Qi, Liu, and Magno}]{huang2024}
Wei Huang, Xingyu Zheng, Xudong Ma, Haotong Qin, Chengtao Lv, Hong Chen, Jie Luo, Xiaojuan Qi, Xianglong Liu, and Michele Magno. 2024.
\newblock \href {https://doi.org/10.48550/arXiv.2404.14047} {An empirical study of llama3 quantization: From llms to mllms}.
\newblock \emph{arXiv preprint arXiv:2404.14047}.

\bibitem[{Jain et~al.(2005)Jain, Nandakumar, and Ross}]{jain2005score}
Anil Jain, Karthik Nandakumar, and Arun Ross. 2005.
\newblock \href {https://doi.org/10.1016/j.patcog.2005.01.012} {Score normalization in multimodal biometric systems}.
\newblock \emph{Pattern Recognition}, 38(12):2270--2285.

\bibitem[{Ji et~al.(2024)Ji, Chen, Jin, Xu, Hua, and Zhang}]{ji2024}
Jianchao Ji, Yutong Chen, Mingyu Jin, Wujiang Xu, Wenyue Hua, and Yongfeng Zhang. 2024.
\newblock \href {https://doi.org/10.48550/arXiv.2406.04428} {Moralbench: Moral evaluation of llms}.
\newblock \emph{arXiv preprint arXiv:2406.04428}.

\bibitem[{Kang and Choi(2023)}]{kang2023}
Cheongwoong Kang and Jaesik Choi. 2023.
\newblock \href {https://doi.org/10.18653/v1/2023.findings-emnlp.518} {Impact of co-occurrence on factual knowledge of large language models}.
\newblock In \emph{Findings of the Association for Computational Linguistics: EMNLP 2023}, pages 7721--7735, Singapore. Association for Computational Linguistics.

\bibitem[{Kappal(2019)}]{kappal2019data}
S.~Kappal. 2019.
\newblock Data normalization using median median absolute deviation mmad based z-score for robust predictions vs. min--max normalization.
\newblock \emph{London Journal of Research in Science: Natural and Formal}, 19(4):39--44.

\bibitem[{Kostova and Radoynovska(2008)}]{kostova2008word}
Zdravka Kostova and Blagovesta Radoynovska. 2008.
\newblock \href {https://zdravka-kostova.com/free/Kostova,%20Radoynovska,%202008.pdf} {Word association test for studying conceptual structures of teachers and students}.
\newblock \emph{Bulgarian Journal of Science and Education Policy (BJSEP)}, 2(2):209--231.

\bibitem[{Liu et~al.(2022)Liu, Cohn, Deyne, and Frermann}]{liu-etal-2022-wax}
Chunhua Liu, Trevor Cohn, Simon~De Deyne, and Lea Frermann. 2022.
\newblock \href {https://doi.org/10.18653/v1/2022.aacl-main.9} {{WAX}: A new dataset for word association e{X}planations}.
\newblock In \emph{Proceedings of the 2nd Conference of the Asia-Pacific Chapter of the Association for Computational Linguistics and the 12th International Joint Conference on Natural Language Processing (Volume 1: Long Papers)}, pages 106--120, Online only. Association for Computational Linguistics.

\bibitem[{Liu et~al.(2021)Liu, Cohn, and Frermann}]{liu-etal-2021-commonsense}
Chunhua Liu, Trevor Cohn, and Lea Frermann. 2021.
\newblock \href {https://doi.org/10.18653/v1/2021.conll-1.38} {Commonsense knowledge in word associations and {C}oncept{N}et}.
\newblock In \emph{Proceedings of the 25th Conference on Computational Natural Language Learning}, pages 481--495, Online. Association for Computational Linguistics.

\bibitem[{Lowe(1997)}]{lowe1997}
Will Lowe. 1997.
\newblock Meaning and the mental lexicon.
\newblock In \emph{Proceedings of the 15th International Joint Conference on Artificial Intelligence (IJCAI)}, pages 1092--1097.

\bibitem[{Mikolov et~al.(2013)Mikolov, Sutskever, Chen, Corrado, and Dean}]{mikolov2013}
Tomas Mikolov, Ilya Sutskever, Kai Chen, Greg~S Corrado, and Jeff Dean. 2013.
\newblock \href {https://proceedings.neurips.cc/paper_files/paper/2013/file/9aa42b31882ec039965f3c4923ce901b-Paper.pdf} {Distributed representations of words and phrases and their compositionality}.
\newblock In \emph{Advances in Neural Information Processing Systems}, volume~26. Curran Associates, Inc.

\bibitem[{Nam et~al.(2024)Nam, Macvean, Hellendoorn, Vasilescu, and Myers}]{nam2024}
Daye Nam, Andrew Macvean, Vincent Hellendoorn, Bogdan Vasilescu, and Brad Myers. 2024.
\newblock \href {https://doi.org/10.1145/3597503.3639187} {Using an llm to help with code understanding}.
\newblock In \emph{ICSE '24: Proceedings of the IEEE/ACM 46th International Conference on Software Engineering}, New York, NY, USA. Association for Computing Machinery.

\bibitem[{Ramezani and Xu(2023)}]{ramezani-2023-knowledge}
Aida Ramezani and Yang Xu. 2023.
\newblock \href {https://doi.org/10.18653/v1/2023.acl-long.26} {Knowledge of cultural moral norms in large language models}.
\newblock In \emph{Proceedings of the 61st Annual Meeting of the Association for Computational Linguistics (Volume 1: Long Papers)}, pages 428--446, Toronto, Canada. Association for Computational Linguistics.

\bibitem[{Ramezani and Xu(2024)}]{Ramezani2024}
Aida Ramezani and Yang Xu. 2024.
\newblock Moral association graph: A cognitive model for moral inference.
\newblock In \emph{Proceedings of the Annual Meeting of the Cognitive Science Society}, volume~46.

\bibitem[{Santurkar et~al.(2023)Santurkar, Durmus, Ladhak, Lee, Liang, and Hashimoto}]{Santurkar2023refelectingopinions}
Shibani Santurkar, Esin Durmus, Faisal Ladhak, Cinoo Lee, Percy Liang, and Tatsunori Hashimoto. 2023.
\newblock Whose opinions do language models reflect?
\newblock In \emph{Proceedings of the 40th International Conference on Machine Learning}, ICML'23. JMLR.org.

\bibitem[{Scherrer et~al.(2023)Scherrer, Shi, Feder, and Blei}]{Scherrer2023}
Nino Scherrer, Claudia Shi, Amir Feder, and David Blei. 2023.
\newblock Evaluating the moral beliefs encoded in llms.
\newblock In \emph{Advances in Neural Information Processing Systems}, volume~36, pages 51778--51809.

\bibitem[{Shin et~al.(2018)Shin, Suh, Eom, and Kim}]{shin2018happiness}
Ji-eun Shin, Eunkook~M. Suh, Kimin Eom, and Heejung~S. Kim. 2018.
\newblock \href {https://doi.org/10.1007/s10902-016-9836-8} {What does “happiness” prompt in your mind? culture, word choice, and experienced happiness}.
\newblock \emph{Journal of Happiness Studies}, 19:649--662.

\bibitem[{Son et~al.(2014)Son, Do, Kim, Cho, Suwonsichon, and Valentin}]{son2014understanding}
Jung-Soo Son, Vinh~Bao Do, Kwang-Ok Kim, Mi~Sook Cho, Thongchai Suwonsichon, and Dominique Valentin. 2014.
\newblock \href {https://doi.org/10.1016/j.foodqual.2013.07.001} {Understanding the effect of culture on food representations using word associations: The case of “rice” and “good rice”}.
\newblock \emph{Food Quality and Preference}, 31:38--48.

\bibitem[{Taxidou and Fischer(2014)}]{taxidou2014}
Io~Taxidou and Peter~M. Fischer. 2014.
\newblock \href {https://doi.org/10.1145/2567948.2580050} {Online analysis of information diffusion in twitter}.
\newblock In \emph{Proceedings of the 23rd International Conference on World Wide Web (WWW '14 Companion)}, pages 1313--1318, New York, NY, USA. Association for Computing Machinery.

\bibitem[{Tjuatja et~al.(2024)Tjuatja, Chen, Wu, Talwalkar, and Neubig}]{tjuatja2024}
Lindia Tjuatja, Valerie Chen, Tongshuang Wu, Ameet Talwalkar, and Graham Neubig. 2024.
\newblock \href {https://doi.org/10.1162/tacl\_a\_00685} {Do llms exhibit human-like response biases? a case study in survey design}.
\newblock \emph{Transactions of the Association for Computational Linguistics}, 12:1011--1026.

\bibitem[{Topkev(2024)}]{topkev2024}
Ahmed Topkev. 2024.
\newblock \href {https://doi.org/10.1007/978-3-031-49519-9_6} {\emph{Framing Religion}}, pages 185--284.
\newblock Springer Nature Switzerland, Cham.

\bibitem[{Van~Rensbergen et~al.(2015)Van~Rensbergen, Storms, and De~Deyne}]{vanrensbergen2015}
Bram Van~Rensbergen, Gert Storms, and Simon De~Deyne. 2015.
\newblock \href {https://doi.org/10.3758/s13423-015-0832-5} {Examining assortativity in the mental lexicon: Evidence from word associations}.
\newblock \emph{Psychonomic Bulletin \& Review}, 22:1717--1724.

\bibitem[{Walker(2006)}]{walker2006}
David~A. Walker. 2006.
\newblock \href {https://doi.org/10.56801/10.56801/v5.i.261} {A comparison of the spearman-brown and flanagan-rulon formulas for split half reliability under various variance parameter conditions}.
\newblock \emph{Archives}, 5(2).

\bibitem[{Warriner et~al.(2013)Warriner, Kuperman, and Brysbaert}]{warriner2013norms}
Amy~Beth Warriner, Victor Kuperman, and Marc Brysbaert. 2013.
\newblock Norms of valence, arousal, and dominance for 13,915 english lemmas.
\newblock \emph{Behavior Research Methods}, 45(4):1191--1207.

\bibitem[{Zheng et~al.(2023)Zheng, Chiang, Sheng, Zhuang, Wu, Zhuang, Lin, Li, Li, Xing, Zhang, Gonzalez, and Stoica}]{zheng2023}
Lianmin Zheng, Wei-Lin Chiang, Ying Sheng, Siyuan Zhuang, Zhanghao Wu, Yonghao Zhuang, Zi~Lin, Zhuohan Li, Dacheng Li, Eric Xing, Hao Zhang, Joseph~E Gonzalez, and Ion Stoica. 2023.
\newblock \href {https://proceedings.neurips.cc/paper_files/paper/2023/file/91f18a1287b398d378ef22505bf41832-Paper-Datasets_and_Benchmarks.pdf} {Judging llm-as-a-judge with mt-bench and chatbot arena}.
\newblock In \emph{Advances in Neural Information Processing Systems}, volume~36, pages 46595--46623. Curran Associates, Inc.

\bibitem[{Zhou et~al.(2003)Zhou, Bousquet, Lal, Weston, and Sch{\"o}lkopf}]{Zhou2003}
Dengyong Zhou, Olivier Bousquet, Thomas Lal, Jason Weston, and Bernhard Sch{\"o}lkopf. 2003.
\newblock Learning with local and global consistency.
\newblock In \emph{Advances in Neural Information Processing Systems 16 (NeurIPS 2003)}. MIT Press.

\end{thebibliography}
\clearpage
\appendix
% \onecolumn
\section{Word Association Test Instructions}
\label{sec:appendix}
We used the following prompt to generate \wallm.\\
\\
\noindent\textbf{System Prompt:}
\begin{itemize}
    \item[] \textbf{Background:} On average, an adult knows about 40,000 words, but what do these words mean to people like you and me? You can help scientists understand how meaning is organized in our mental dictionary by playing the game of word associations. This game is easy: Just give the first three words that come to mind for a given cue word.
    
    \item[] \textbf{Output Format:} Output your response in the following format: \\ 
    % \begin{verbatim}
    response1, response2, response3 \\ 
    % \end{verbatim}
    Do not provide any additional context or explanations. Just the words as comma-separated values.
\end{itemize}

\noindent\textbf{User Prompt:} Cue word: \texttt{\{keyword\}}
\newline

The fixed system prompt positions the model as a human participant in a psychology experiment, requesting three word associations for a given cue word, formatted as comma-separated values without additional context. The exact same system prompt has been used to collecting human responses for \href{https://smallworldofwords.org/en}{\wahuman}. The \{keyword\} will be replaced with actual cue words when generating word associations, and each cue will be prompted 100 times.

\section{\wahuman and \wallm Reliability Test}
\label{sec:Reliability Test}
\begin{figure}[t]
    \centering
    \includegraphics[width=0.45\textwidth]{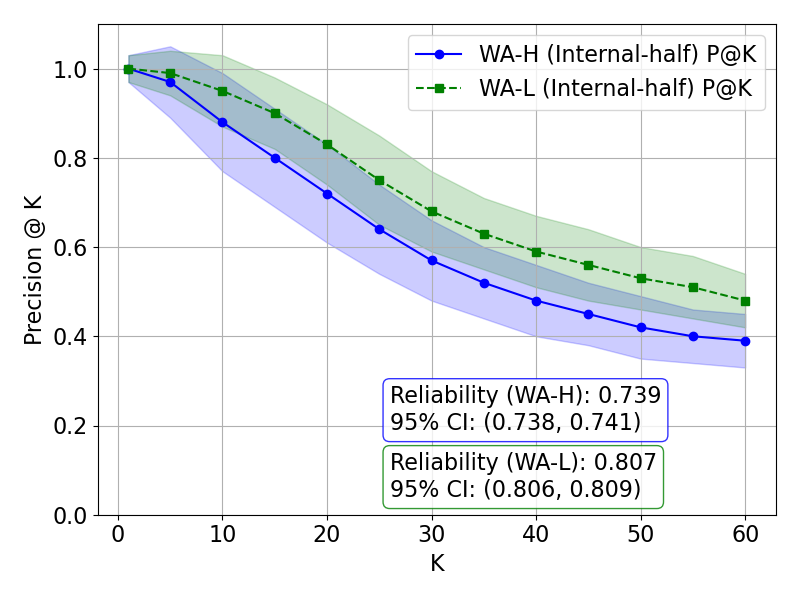}
    \caption{Precision@K for \wahuman and \wallm associations.}
    \label{fig: r_h_W}
\end{figure}

Figure~\ref{fig: r_h_W} presents reliability test for \wallm and \wahuman using the the Precision@K.

\wahuman refers to word associations produced by human participants, as detailed in Section \ref{sec: dataset}. The figure compares precision@K for each internal half. Each line shows precision at different K values, with shaded regions representing standard deviation over 50 runs. Reliability values are noted
% \clearpage

\section{Graph Statistics}
Table~\ref{tab:Graph Statistic} presents the overall graph statistics of \wahuman and \wallm. Both graphs were prompted with the same 12,216 cue words. 

Compared to \wahuman, \wallm has fewer edges, lower density, and lower average connectivity, but exhibits a slightly higher local clustering coefficient and a larger diameter, indicating more localized subgraph connections.
\label{sec:Graph Statistic}
\begin{table}[t]
\centering
\begin{tabular}{l|c|c}
\toprule
\textbf{} & \textbf{\wahuman} & \textbf{\wallm} \\ \midrule 
\textbf{Nodes} & 12,216 & 12,216 \\ 
\textbf{Edges Number} & 963,043 & 502,174 \\ 
\textbf{Density} & 0.013 & 0.007 \\ 
\textbf{Local Cluster} & 0.12 & 0.15 \\ 
\textbf{Max Connectivity} & 221 & 208 \\ 
\textbf{Min Connectivity} & 48 & 10 \\ 
\textbf{AVG Connectivity} & 114 & 77 \\ 
\textbf{SD Connectivity} & 21 & 23 \\ 
\textbf{Diameter} & 3 & 4 \\ \bottomrule
\end{tabular}
\caption{A statistical overview of the global word association graphs in \wahuman and \wallm.}
\label{tab:Graph Statistic}
\end{table}

\section{Optimizing Alpha}
\label{sec:appendix2}
\begin{figure}[h]
    \centering
    \includegraphics[width=0.45\textwidth]{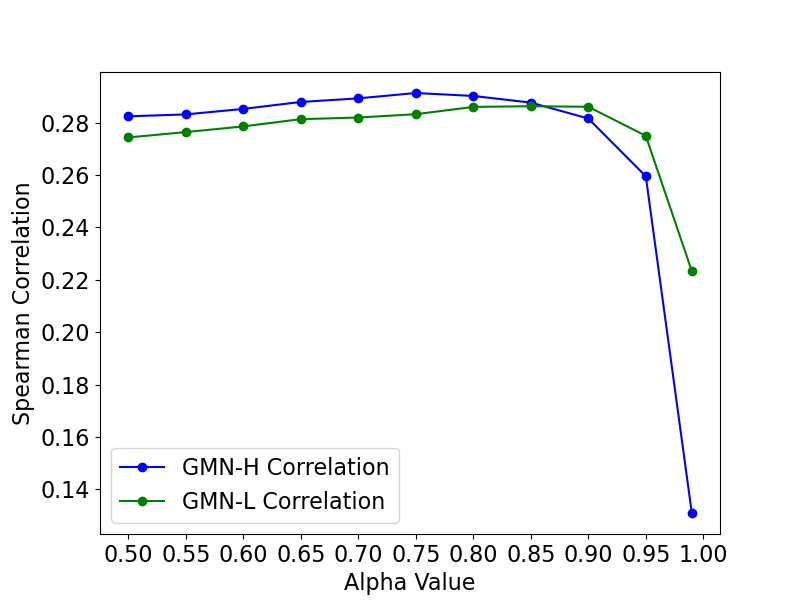}
    \caption{The Spearman correlation between the eMFD and the propagated values for various values of $\alpha$.}
    \label{fig: varying alpha}
\end{figure}

Figure~\ref{fig: varying alpha} shows how the Spearman correlation varies with different $\alpha$ values for both \swow and \llama. 

The \llama correlation reaches its peak at alpha = 0.75, while the \swow correlation peaks at alpha = 0.9. We used these respective optimal values in Section~\ref{sec:gmsn} to propagate the moral values.
\section{Ranking Values}

\label{sec:Ranking Values}
We present the top-ranked positive and negative words that we used, as well as words with different polarity in the Section~\ref{sec:evaluation} supplemented with their overall morality score and dimensions. 

The morality score is calculated as the sum of scores across five dimensions after propagation. Due to differences in word association responses between LLMs and humans, the values produced may not be directly comparable. To address this, we applied median absolute deviation (MAD) normalization post-aggregation to the sum scores for both LLMs and humans. This helps ensure consistency in comparisons across potentially skewed distributions and mitigating outliers, while still preserving the internal structure of the data.\cite{jain2005score, kappal2019data}. 

The moral dimension of a concept is the one with the highest score among the five dimensions. Denoting the dominant dimensions as 1: Care, 2: Fairness, 3: Loyalty, 4: Authority, 5: Sanctity.

\subsection{Top Negative}
\swow: disgusting(5): -28, traitor(3): -27, vomit(5): -27, hurt(1): -26, dirty(5): -26, pain(1): -25, bad(5): -25, thief(2): -24, gross(5): -24, sick(5): -24 \newline

\noindent \llama:  betrayal(2): -43, prejudice(2): -38, cheating(2): -37, disgusting(2): -36, discrimination(2): -33,  dishonest(2): -32, deception(2): -31, dishonesty(2): -30, racism(2): -30, infidelity(3): -28

\subsection{Top Positive}
\swow: church(5): 62.03, religion(5): 52.71, God(5): 47.78, priest(5): 37.43, holy(5): 34.74, religious(5): 34.04, catholic(5): 33.01, kind(1): 29.72, caring(1): 26.04, worship(4)(5): 25.72\newline

\noindent \llama:  church(5): 52, kindness(1): 41, religion(5): 40, priest(5): 36, prayer(5): 34, bible(5): 34, faith(5): 34, family(3): 33,  compassion(1): 32, holy(5): 30

\subsection{Difference}

Table~\ref{tab:divergent_concepts_scores} presents the concepts that we used in Table~\ref{tab:comparison_table} (column Different), along with their dominant moral dimensions (using \swow as the standard) and propagated moral scores from \swow and \llama.

\begin{table}[h]
\centering

\setlength{\tabcolsep}{1.8pt}
\begin{tabular}{c|c|c}
\toprule
\textbf{Word (Dimension)} & \textbf{\swow} & \textbf{\llama} \\
\midrule
Abortion (1,4) & -0.45 & 1.5 \\
Immigrant (4) & -0.62 & 1.1 \\
Politician (2) & -6.6 & 6.5 \\
Capitalist (3,4) & -0.16 & 0.97 \\
Homosexual (4,5) & -0.55 & 1.03 \\
Commercial (2,4,5) & -0.42 & 0.52 \\
Jail (4) & 0.06 & -3.15 \\
Air (4) & 1.09 & -0.73 \\
Plastic (4,5) & 0.19 & -1.25 \\
Soviet (3) & 2.27 & -0.44 \\
Bees (3,4) & 0.23 & -0.82 \\
Snob (4) & 1.15 & -0.32 \\
\bottomrule
\end{tabular}
\caption{Comparison of concepts with divergent moral values from \swow and \llama.\label{tab:divergent_concepts_scores}}
\end{table}

\section{Response Analysis}
\label{sec:response_analysis}
For cue words in the Table \ref{tab:comparison_table}, we provide the detailed associations to understand how their moral values are being captured by \swow and \llama. 
We examine (a) the top frequent responses for each cue word and in both \swow and \llama; and (b) ``top unique response'': a response that appears in one graph (\llama or \swow) but does not appear in the other.

\subsection{Negative Response Analysis}
\label{sec:negative_response_analysis}
% \label{sec:response analysis}
Table~\ref{tab:negative_response} presents the associations for the representative top negative moral concepts in Table \ref{tab:comparison_table} that we manually selected.

\begin{table*}[!t]
    \centering
    % \small
    % \resizebox{\textwidth}{!}{%
    \begin{tabular}{|c|c c|c c|}
        \hline
        \textbf{Cue Word} & \multicolumn{2}{c|}{\textbf{Top Response}} & \multicolumn{2}{c|}{\textbf{Top Unique Response}} \\ \hline
        & \textbf{\swow} & \textbf{\llama} & \textbf{\swow} & \textbf{\llama} \\ \hline
        \textbf{prejudice} 
        & pride & bias 
        & pride & stereotypes \\
        & racism & racism 
        & black & biases \\
        & black & discrimination 
        & race & stereotyping \\
        & race & bigotry 
        & racist & bigoted \\ \hline
        
        \textbf{racism}
        & black & prejudice 
        & black & inequality \\
        & white & discrimination 
        & white & segregation \\
        & bad & bigotry 
        & bad & equality \\
        & prejudice & inequality 
        & bigot & pain \\ \hline
        
        \textbf{discrimination}
        & racism & prejudice 
        & race & stereotypes \\
        & race & racism 
        & racist & stereotyping \\
        & prejudice & bias 
        & sexism & equality \\
        & unfair & inequality 
        & gender & prejudices \\ \hline
        
        \textbf{vomit}
        & puke & nausea 
        & gross & stomachache \\
        & sick & sickness 
        & spew & queasy \\
        & gross & stomach 
        & smell & hangover \\
        & barf & stomachache 
        & green & poisoning \\ \hline
        
        \textbf{gross}
        & disgusting & disgusting 
        & fat & nauseating \\
        & nasty & vomit 
        & net & disgusted \\
        & ugly & nauseating 
        & large & queasy \\
        & fat & revolting 
        & yuck & nausea \\ \hline
    \end{tabular}%
    % }
    \caption{Comparison of the top 4 responses and top 4 unique responses between \swow and \llama for selected cue words in top negative and divergent concepts, ranked based on frequency.}
    \label{tab:negative_response}
\end{table*}

% \clearpage

\subsection{Positive Response Analysis}
\label{sec:positive response analysis}

Table~\ref{tab:positive response analysis} presents the associations for the representative top negative moral concepts in Table~\ref{tab:comparison_table} that we manually selected.

\begin{table*}[htbp]
    \centering
    % \small
    %\setlength{\extrarowheight}{-1pt}
    % \resizebox{\textwidth}{!}{%
    \begin{tabular}{|c|c c|c c|}
        \hline
        \textbf{Cue Word} & \multicolumn{2}{c|}{\textbf{Top Response}} & \multicolumn{2}{c|}{\textbf{Top Unique Response}} \\ \hline
        & \textbf{\swow} & \textbf{\llama} & \textbf{\swow} & \textbf{\llama} \\ \hline
        \textbf{kind} 
        & nice & gentle & type & nurturing \\
        & type & caring & sort & soft \\
        & gentle & friendly & happy & charitable \\
        & sweet & compassionate & person & warmth \\ \hline
        
        \textbf{caring}
        & love & nurturing & sharing & supportive \\
        & loving & loving & nice & motherly \\
        & kind & kind & giving & selfless \\
        & sharing & compassionate & sweet & emotional \\ \hline
        
        \textbf{church}
        & steeple & altar & catholic & altar \\
        & religion & priest & synagogue & minister \\
        & God & sunday & stone & baptism \\
        & priest & pews & school & service \\ \hline
        
        \textbf{priest}
        & church & church & father & altar \\
        & catholic & clergy & black & clergyman \\
        & father & altar & vicar & chapel \\
        & religion & minister & pedophile & vatican \\ \hline
        
        \textbf{religion}
        & God & church & cross & beliefs \\
        & church & faith & war & rituals \\
        & faith & God & atheism & scripture \\
        & Christianity & spirituality & fear & churches \\ \hline
    \end{tabular}%
    % }
    \caption{Comparison of the top 4 responses and top 4 unique responses between \swow and \llama for selected cue words in top positive concepts, ranked based on frequency.}
    \label{tab:positive response analysis}
\end{table*}
\clearpage
\onecolumn
\section{Quantitative analysis of graph property}
\label{sec: subgraph}
\begin{figure}[h]
    \centering 
    \includegraphics[width=\columnwidth]{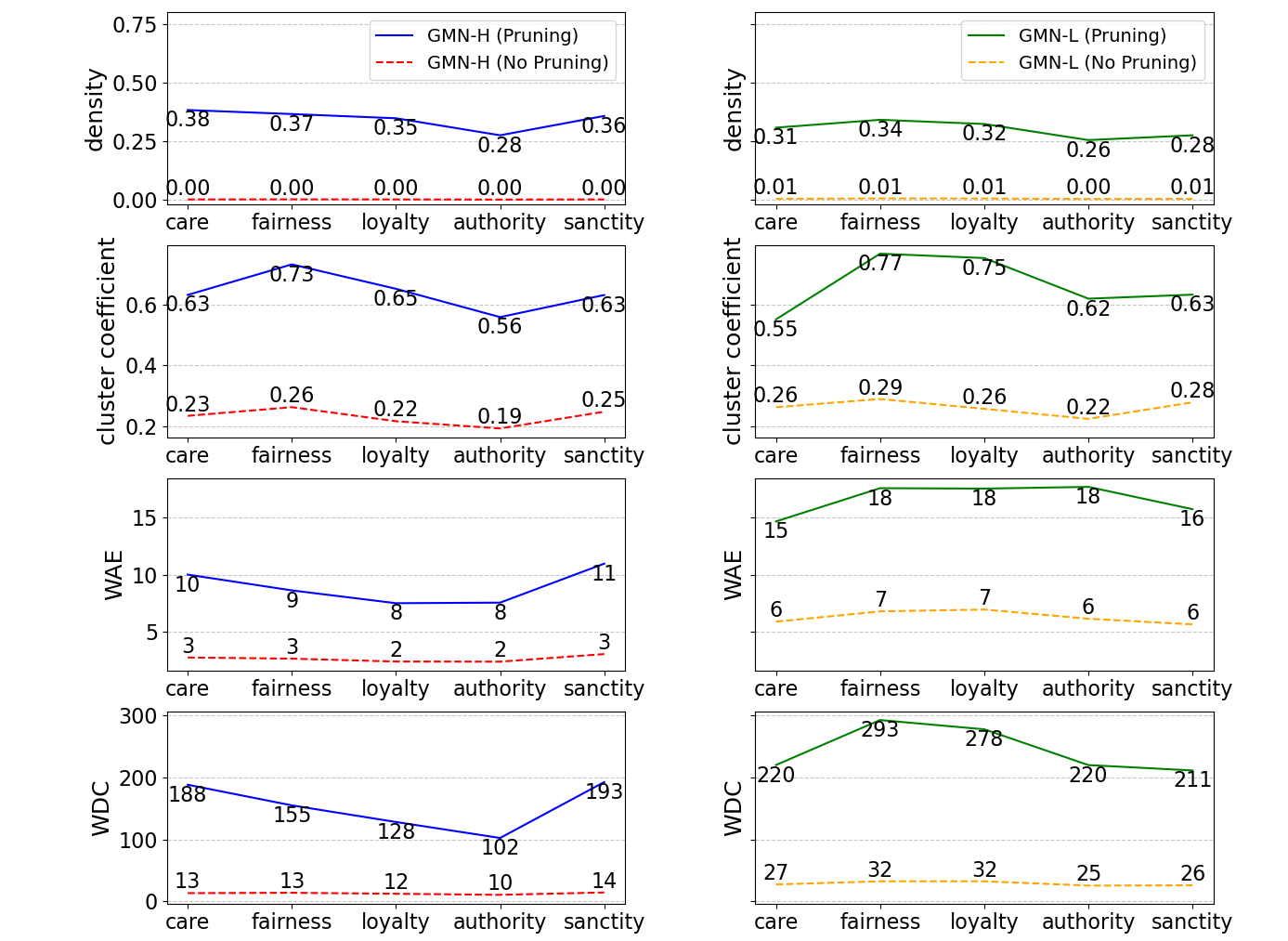}
    \caption{Quantitative analysis of graph properties—density, local clustering coefficient (clustering coefficient), weighted average edge (WAE), and weighted degree centrality (WDC)—was conducted across moral dimensions for both \swow and \llama. Results are presented for pruned and non-pruned subgraphs, highlighting the effects of pruning on propagation efficiency and network density. In pruned subgraphs, we keep only the top 50 negative cues based on each dimension in the graph. In non-pruned subgraphs, the subgraph contains not only the top 50 negative cues but also each cue's corresponding responses. WAE represents the average edge connection weight between any two connected nodes in a graph, with higher WAE indicating a greater potential for moral information transfer during propagation.}
    \label{fig:Pruning}
\end{figure}

Figure~\ref{fig:Pruning} presents detailed the graph analysis we used in Section~\ref{sec:evaluation}.

\end{document}